\documentclass[letterpaper, 10 pt, conference]{ieeeconf}  

\usepackage{graphicx}
\usepackage{multirow}
\usepackage{textcomp}
\usepackage{amssymb,amsmath}
\usepackage{hyperref}
\usepackage{booktabs}
\usepackage{caption}
\usepackage{subcaption}
\usepackage{algorithm2e}
\usepackage{siunitx}
\usepackage{physics}
\AtBeginDocument{\RenewCommandCopy\qty\SI}

\IEEEoverridecommandlockouts                              

\title{\LARGE \bf Sample-efficient Real-time Planning with Curiosity Cross-Entropy Method and Contrastive Learning}

\author{Mostafa Kotb$^{1,2, \ast}$, Cornelius Weber$^{1}$, and Stefan Wermter$^{1}$%
\thanks{$^{1}$The authors are with the Knowledge Technology Group, Department of Informatics, Universität Hamburg, 22527 Hamburg, Germany. E-mail: $\{$mostafa.kotb, cornelius.weber, stefan.wermter$\}$@uni-hamburg.de.}
\thanks{$^{2}$Mathematics Department, Faculty of Science, Aswan University, 81528 Aswan, Egypt.}%
\thanks{$^{\ast}$Corresponding author, Email: mostafa.kotb@uni-hamburg.de.}
}

\begin{document}

\maketitle
\thispagestyle{empty}
\pagestyle{empty}

\begin{abstract}

Model-based reinforcement learning (MBRL) with real-time planning has shown great potential in locomotion and manipulation control tasks. However, the existing planning methods, such as the Cross-Entropy Method (CEM), do not scale well to complex high-dimensional environments. One of the key reasons for underperformance is the lack of exploration, as these planning methods only aim to maximize the cumulative extrinsic reward over the planning horizon. Furthermore, planning inside the compact latent space in the absence of observations makes it challenging to use curiosity-based intrinsic motivation. We propose Curiosity CEM (CCEM), an improved version of the CEM algorithm for encouraging exploration via curiosity. Our proposed method maximizes the sum of state-action $Q$ values over the planning horizon, in which these $Q$ values estimate the future extrinsic and intrinsic reward, hence encouraging to reach novel observations. In addition, our model uses contrastive representation learning to efficiently learn latent representations. Experiments on image-based continuous control tasks from the DeepMind Control suite show that CCEM is by a large margin more sample-efficient than previous MBRL algorithms and compares favorably with the best model-free RL methods.

\end{abstract}
\section{INTRODUCTION}

Model-based RL (MBRL) improves sample efficiency by learning a dynamics model in latent space, then either utilizes the learned model directly for real-time (online) planning \cite{planet}, \cite{tdmpc} or optimizes a policy inside imagined trajectories (i.e., background planning) \cite{dreamer}. MBRL has shown outstanding successes in complex discrete environments, such as defeating human world champions in chess \cite{chess} and Go \cite{go}. However, in continuous control tasks, planning methods such as the Cross-Entropy Method (CEM) \cite{cem} do not scale well with the increasing complexity in environments. The way of planning by randomly generating action sequences and then executing the first action in the sequence with the highest expected reward, is inefficient in complex high-dimensional environments \cite{mbrl1}, \cite{disadvantage1}. 

Furthermore, CEM lacks exploration as it only aims to maximize the extrinsic reward of the sampled action sequences, therefore it might fail in sparse reward settings and in hard-to-explore environments with high-dimensional state and action spaces. Consequently, no MBRL algorithm has yet achieved the asymptotic performance as the best model-free RL algorithm on image-based continuous tasks \cite{mbrl2}. 

\begin{figure}
    \centering
    \includegraphics[width=0.53\textwidth]{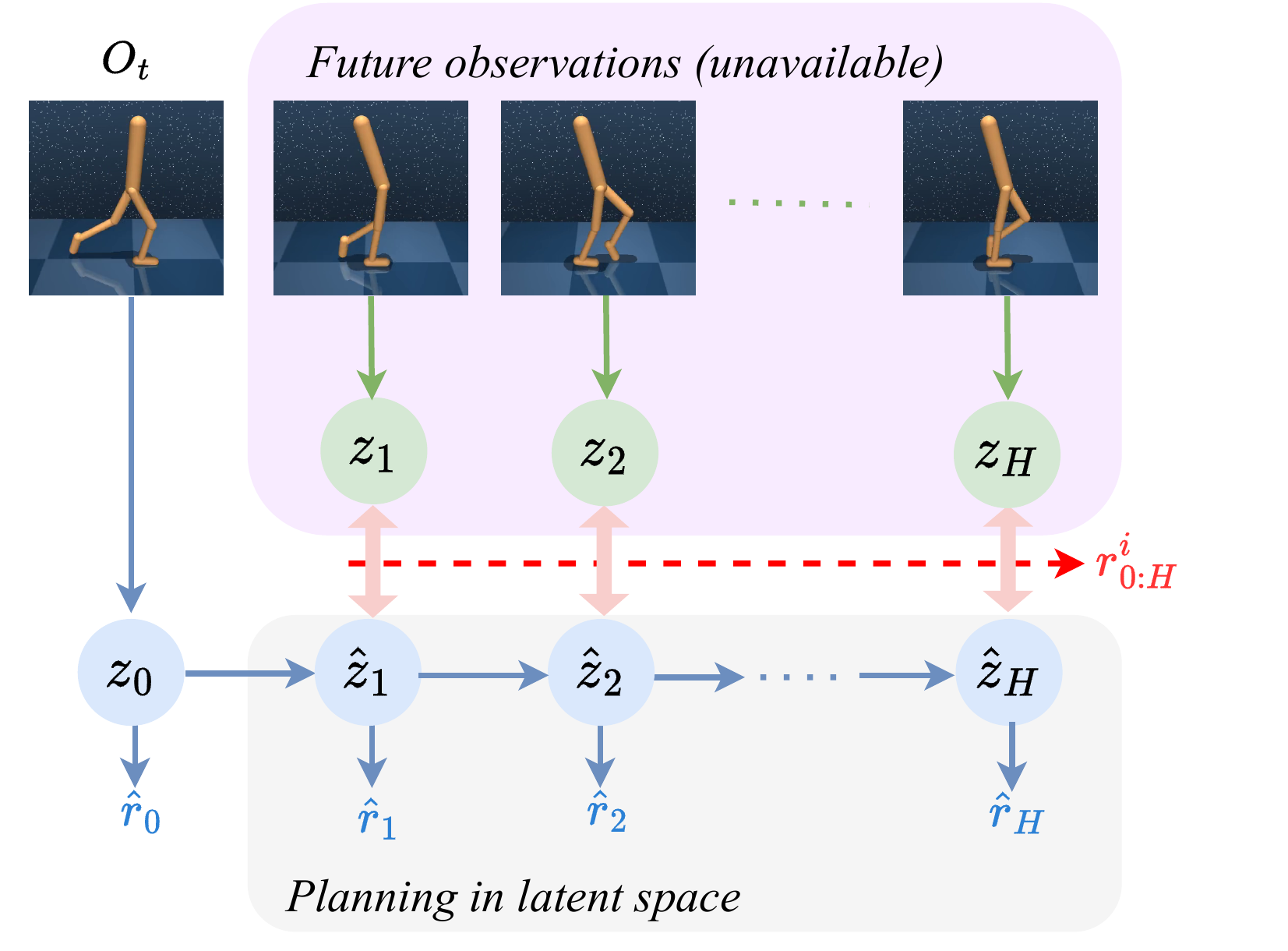}
    \caption{\emph{The challenge of using intrinsic reward during planning:} At every time step, the current observation $o_t$ is encoded into latent state $z_0$. Then, the planner, with the help of the learned latent dynamics model plans a trajectory of length $H$ inside the latent space by predicting latent states $\hat{z}_{1:H}$ and extrinsic rewards $\hat{r}_{0:H}$. Future observations are not available during planning, therefore the intrinsic rewards $r^i_{0:H}$ cannot be estimated as the prediction error between the actual latent states $z_{1:H}$ and the predicted latent states $\hat{z}_{1:H}$ \textit{(as indicated by the dashed red arrow)}.}
    \label{fig:0}
\end{figure}

An effective approach to improve exploration is to use curiosity-based intrinsic reward as the prediction error of the next latent state (i.e., prediction-based exploration) \cite{icm}, \cite{intrinsic1}, \cite{intrinsic2}, which encourages reaching novel states. 
Curiosity-based intrinsic motivation has been used extensively with model-free RL algorithms \cite{icm}, \cite{intrinsic2}, \cite{vime}, \cite{intrinsic3}, but is still rarely used with MBRL \cite{plan2explore}. Unfortunately, it is technically challenging to use such an intrinsic reward with MBRL planning methods, especially real-time planning (see Fig. \ref{fig:0}), as the ground-truth future observations are unavailable during planning, and hence the prediction error cannot be estimated. 

A solution proposed in Plan2Explore \cite{plan2explore} is to compute the intrinsic reward as the disagreement in the predicted next latent state from an ensemble of forward dynamics models. However, training an ensemble of forward dynamics models in addition to a latent dynamics model is computationally intensive and requires careful balancing of the heterogeneity of the population.

In this paper, to alleviate the aforementioned challenges with real-time planning, we propose \textit{Curiosity Cross-Entropy Method (CCEM)}, an improved version of CEM for encouraging exploration via curiosity. We take a different route from Plan2Explore: instead of estimating the intrinsic reward \textit{online} during planning using an ensemble of forward dynamics models, our proposed method estimates the intrinsic reward \textit{offline} during training using an \textit{Intrinsic Curiosity Module} \cite{icm}. To this end, we train a state-action $Q$ function to estimate future extrinsic and intrinsic reward. During planning, the proposed Curiosity CEM maximizes the sum of these $Q$ values over the planning horizon, hence encouraging to reach novel states. 
To further improve sample efficiency, we use contrastive representation learning by maximizing the temporal mutual information between embeddings of consecutive time steps \cite{temporal1}, \cite{temporal2}, \cite{cody}. We choose TD-MPC \cite{tdmpc} as the model-based RL to evaluate our proposed CCEM and we name it \textit{TD-MPC with CCEM}. 

We evaluate the sample efficiency of our proposed method on six image-based continuous control tasks from the DeepMind Control Suite \cite{dm_suite}. Our proposed method outperforms state-of-the-art model-free RL methods at the 100k environment step, particularly outperforming previous model-based RL algorithms by a large margin, showing its superiority as a real-time planning method. The contributions of our work are as follows:
\begin{itemize}
    \item We propose CCEM, a real-time planning method for encouraging exploration via curiosity.

    \item We demonstrate the robustness of CCEM as a real-time planner by comparing it against two variants of CEM.

    \item We show that TD-MPC with CCEM is more sample efficient than previous MBRL methods.
\end{itemize}

To the best of our knowledge, this is the first time a curiosity-based exploration technique is used to improve the performance of a real-time planning MBRL algorithm.
\section{Related Work}

\subsection{Curiosity-based Exploration}
RL agents are trained by maximizing the cumulative extrinsic reward that is often designed as a dense well-shaped reward \cite{shaping} to facilitate the completion of the task. However, reward shaping requires domain knowledge and human effort, and thus sparse reward tasks are more common in practice at the cost of a slow learning process \cite{sparse}. Curiosity-based exploration has been proposed to help agents explore in sparse reward settings and in complex high-dimensional environments. There have been many techniques introduced, such as \textit{visit-counts} \cite{count1}, \cite{count2}, \cite{count3} which discourages revisiting the same states, and \textit{prediction-based} \cite{icm}, \cite{intrinsic1}, \cite{intrinsic2} which encourages reaching novel states by estimating the intrinsic reward as the prediction error of the next state. To efficiently explore in stochastic environments such as robotics, an ensemble of dynamics models is used and the intrinsic reward is estimated as the disagreement of the ensemble \cite{ensemble1}, \cite{ensemble2}, \cite{ensemble3}, \cite{ensemble4}. A new paradigm introduced in \cite{IRRL} where the reward is generated \textit{internally} using a discriminator that evaluates the novelty of the state.

Prediction-based exploration has shown to be effective and has been used extensively with model-free agents \cite{vime}, \cite{intrinsic3}, \cite{icm}, \cite{intrinsic2} but has been rarely used with model-based agents \cite{plan2explore}. In model-based RL, real-time planning in the latent space in the absence of the ground-truth observations makes it challenging to estimate the intrinsic reward as the prediction error of next state. To overcome this challenge, we propose to compute the intrinsic reward offline during training using Intrinsic Curiosity Module \cite{icm}, as the ground-truth observations are available. Then, we train a state-action $Q$ value function to estimate future extrinsic and intrinsic reward. During planning, we maximize the sum of $Q$ values over the planning horizon. 

\subsection{Contrastive Representation Learning}
Recently, contrastive learning  \cite{cl1} has proven effective in learning latent representations and led to improve the sample efficiency of vision-based RL agents. Contrastive learning learns latent representations in an unsupervised fashion by minimizing the distance in the latent space between two similar images (i.e., positive pairs), and at the same time maximizing the distance between two dissimilar images (i.e., negative pairs). 

CURL \cite{curl} proposed a contrastive loss between two different data-augmentation of the same observation, while CPC \cite{infonce} and ST-DIM \cite{temporal1} proposed different variations of temporal contrastive loss between two augmented observations separated by small time steps. To decouple representation learning from policy learning, a new unsupervised learning task called Augmented Temporal Contrast was introduced to train the encoder exclusively using a temporal contrastive loss \cite{atc}. The result showed that training the representations in an unsupervised way (i.e. not relying on the environment's reward) is very helpful for multitasking and sparse reward environment. In addition, several contrastive approaches extend model-free RL with a predictive model to help learning temporally consistent representations \cite{cody}, \cite{SPR}, \cite{ccfdm}.

Reconstruction-free model-based RL \cite{dreaming}, \cite{cvrl}, \cite{tpc}, \cite{dreamerpro} learns a world model in a contrastive way without reconstructing the observations. These models succeeded in learning task-related representations in complex observations where task-irrelevant information are presented as distractions. In our work, we use a temporal contrastive loss between the joint representations of an observation and action and the representation of the next observation \cite{cody}.
\begin{figure*}[ht!]
\centering
\includegraphics[width=\textwidth]{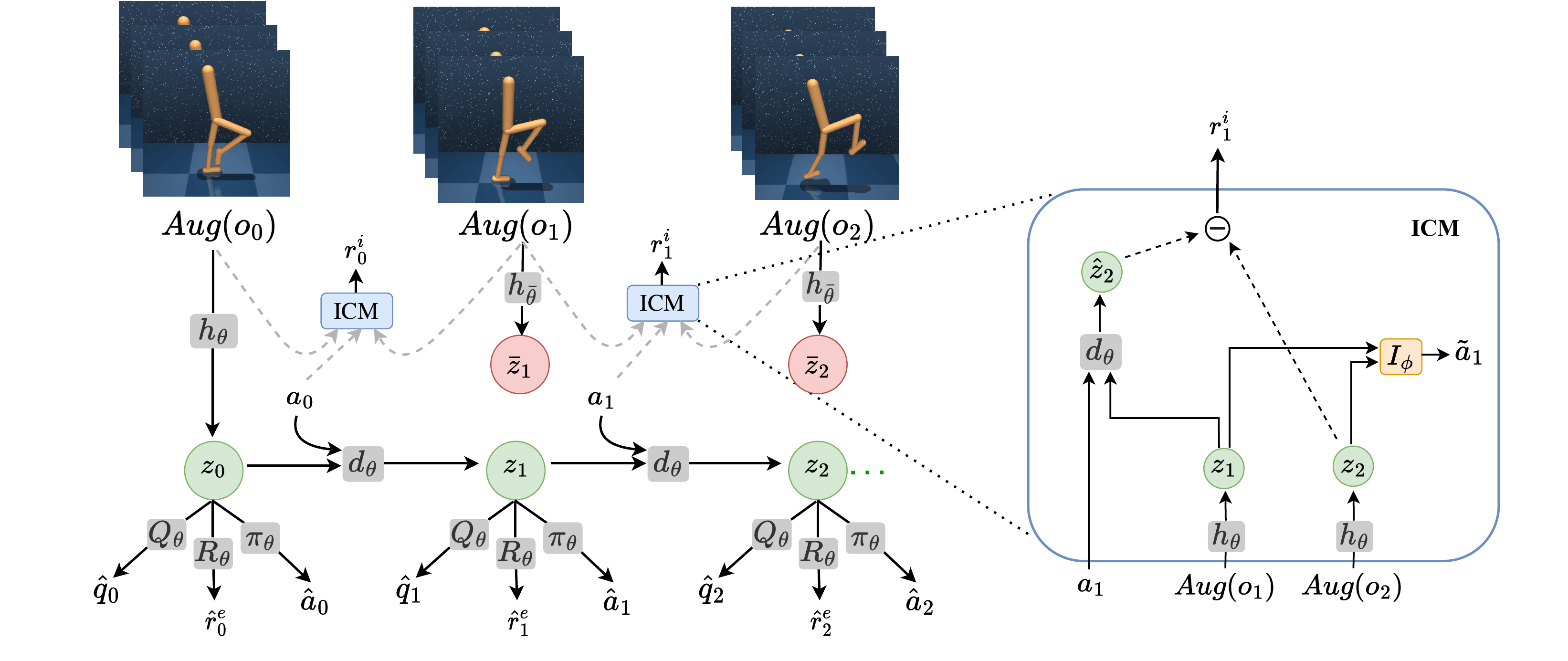}
\caption{\emph{Training:} A trajectory of length $k$ is sampled from the replay buffer. The initial observation $o_0$ is augmented ($Aug(.)$ is $\pm4$ pixel shift augmentation \cite{drq}) and encoded using the online encoder $h_{\theta}$ into latent state $z_0$, and subsequent observations are augmented and encoded using the target encoder $h_{\bar{\theta}}$ which is defined as an exponential moving average of the online encoder into target latent states $\bar{z}_1, \bar{z}_2,...,\bar{z}_k$. TOLD \cite{tdmpc} recurrently predicts latent states $z_1, z_2,...,z_k$, a $Q$-value $\hat{q}_t$, an extrinsic reward $\hat{r}^e_t$, and an action $\hat{a}_t$ for each latent state $z_t$. Intrinsic Curiosity Module (ICM) \cite{icm} is utilized to compute the intrinsic reward $r^i_t$ as the prediction error between the predicted latent state $\hat{z}_{t}$ and the ground-truth ${z}_{t}$.} 
\label{fig:1}
\end{figure*}

\section{Background}
\subsection{Reinforcement Learning from Images with Intrinsic Reward}
We formulate the problem of imaged-based continuous control as an infinite-horizon Markov Decision Process (MDP). An MDP characterized by a tuple $(\mathcal{O}, \mathcal{A}, \mathcal{R}, \mathcal{P}, \gamma)$, where $\mathcal{O}$ is the high-dimensional observation space (RGB image pixels), $\mathcal{A}$ is the continuous action space, $\mathcal{P}:\mathcal{O}\times\mathcal{A}\times\mathcal{O}\mapsto\mathbb{R}_{+}$ is the transition function, $\mathcal{R}:\mathcal{O}\times\mathcal{A}\mapsto\mathbb{R}$ is a reward function (also known as extrinsic reward $r^e_t$), and $\gamma \in [0,1]$ is a discount factor. The goal of RL is to learn a parameterized mapping policy $\Pi_{\theta}:\mathcal{O}\mapsto\mathcal{A}$ that maximizes the expected cumulative reward $\mathbb{E}_{a_t \sim \Pi_{\theta}} [ \sum^{\infty}_{t=0} \gamma^{t}r^e_t]$.

To encourage exploration and avoid the policy from getting stuck in a local minimal, exploration bonuses are given as \textit{intrinsic reward} $r^i_t$. Thus, during training, to encourage reaching novel states, the policy has to maximize the new expected cumulative reward $\mathbb{E}_{a_t \sim \Pi_{\theta}} [ \sum^{\infty}_{t=0} \gamma^{t} (r^e_t + r^i_t)]$.

\subsection{Cross-Entropy Method}
Cross-Entropy Method (CEM) \cite{cem} is a derivative-free optimization technique that has been used with model-based RL as an efficient online planner \cite{planet}, \cite{mbrl1}, \cite{mbrl2}. CEM starts by sampling action sequences of length $H$, where $H$ is the planning horizon, from a time-dependent diagonal Gaussian distribution initialized by zero mean and unit variance $(\mu_{0:H},\sigma_{0:H})$. Then, the sampled sequences are evaluated based on \textit{a scoring function} and the top $k$ candidates are selected. The distribution $\mu$ and $\sigma$ are fitted to the top $k$ candidates and after several iterations of this procedure, the planner returns the mean for the current time step, with $\mu_{t}$ as the best action to be executed. To plan for the next time step, the Gaussian distribution is initialized again to zero mean and unit variance to avoid local optima.

There are three variants of CEM based on three different scoring functions proposed in the literature as follows:

\textit{Sum of rewards} \cite{mbrl2}: Discounted sum of rewards $\sum^{H}_{t=0} \gamma^{t}r(o_t,a_t)$, which defines the original CEM.

\textit{Sum of rewards $+$ terminal value} \cite{terminal_value}: Discounted sum of rewards summed with the estimated value of the terminal state $\sum^{H-1}_{t=0} \gamma^{t}r(o_t,a_t) + \gamma^{H}Q(o_H,a_H)$, which defines CEM with terminal value function.

\textit{Sum of values} \cite{value-sum}: Discounted sum of state-action $Q$ values $\sum^{H}_{t=0} \gamma^{t}Q(o_t,a_t)$, which defines CEM with value summation.

In this paper, we propose \textit{Curiosity CEM}, a fourth variant of CEM for encouraging exploration. The scoring function is the same as the sum of values \cite{value-sum}, except that the $Q$ values are trained to estimate extrinsic and intrinsic reward.

\section{TD-MPC with Curiosity CEM}
 \begin{figure*}[ht]
\centering
     \begin{subfigure}[b]{0.16\textwidth}
         \centering
         \includegraphics[width=\textwidth]{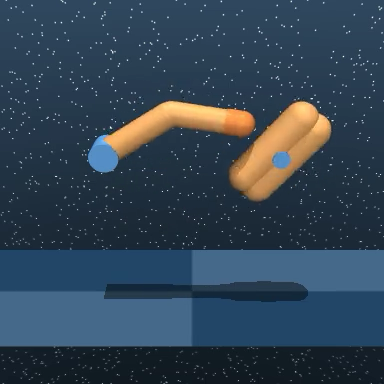}
         \caption{Finger Spin}
     \end{subfigure}
     \begin{subfigure}[b]{0.16\textwidth}
         \centering
         \includegraphics[width=\textwidth]{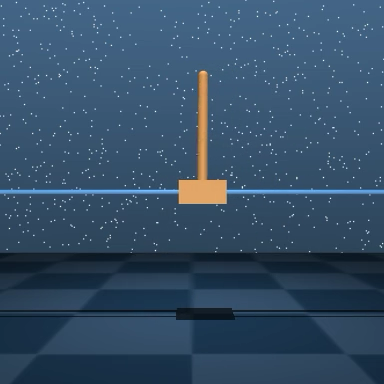}
         \caption{Cartpole Swingup}
     \end{subfigure}
     \begin{subfigure}[b]{0.16\textwidth}
         \centering
         \includegraphics[width=\textwidth]{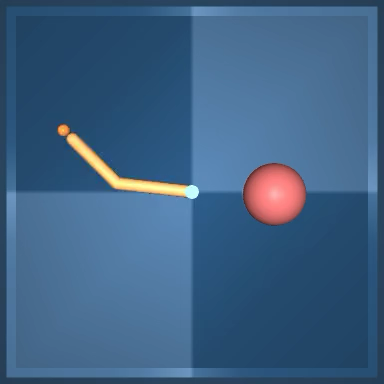}
         \caption{Reacher Easy}
     \end{subfigure}
     \begin{subfigure}[b]{0.16\textwidth}
         \centering
         \includegraphics[width=\textwidth]{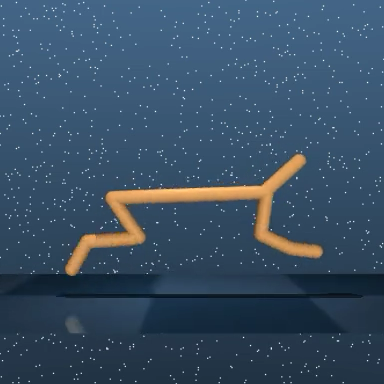}
         \caption{Cheetah Run}
     \end{subfigure}
     \begin{subfigure}[b]{0.16\textwidth}
         \centering
         \includegraphics[width=\textwidth]{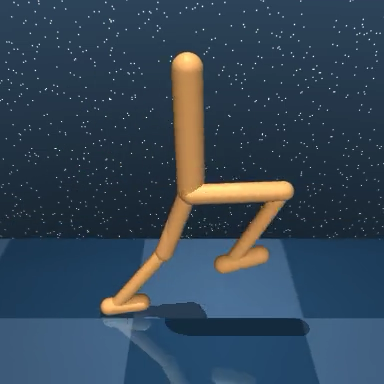}
         \caption{Walker Walk}
     \end{subfigure}
     \begin{subfigure}[b]{0.16\textwidth}
         \centering
         \includegraphics[width=\textwidth]{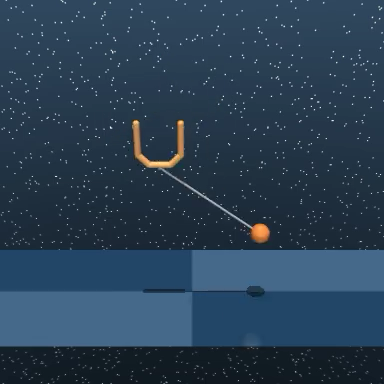}
         \caption{Ball-in-cup Catch}
     \end{subfigure}
\caption{Image-based continuous control tasks from the DeepMind Control Suite. These tasks introduce a diverse set of challenges: sparse reward (e.g., Ball-in-cup, Reacher), complex dynamics (e.g., Walker, Finger, Cheetah), hard exploration (e.g., Walker, Cheetah) due to high action space, and partial observability (e.g., Cartpole as the cart can move out of sight).}
\label{fig:2}
\end{figure*}

We choose Temporal Difference Model Predictive Control (TD-MPC) \cite{tdmpc} as the model-based RL algorithm to test and evaluate our proposed Curiosity CEM (CCEM) method. The original TD-MPC uses CEM with terminal value function as the planning method. In this section, we explain in detail the \textit{training} and \textit{inference} procedures of TD-MPC with CCEM. See Algorithm \ref{alg:training} for training pseudo code.

\subsection{Training}

TD-MPC uses a Task-Oriented Latent Dynamics (TOLD) model which is jointly trained together with a terminal $Q$ value function using temporal difference learning. TOLD consists of five model components (shown as gray shaded squares in Fig.\ref{fig:1}) as follows:
\begin{enumerate}
     \item \textbf{Encoder:} \(z_{t} = h_{\theta}(o_{t})\), encodes a given observation $o_{t}$ into a latent representation $z_{t}$.

     \item \textbf{Latent dynamics:} $z_{t+1} = d_{\theta}(z_{t},a_{t})$, predicts the next latent representation $z_{t+1}$ given $z_{t}$ and action $a_{t}$.

     \item \textbf{Reward:} $\hat{r}^{e}_{t} = R_{\theta}(z_{t},a_{t})$, predicts extrinsic reward $\hat{r}^{e}_{t}$ given $z_{t}$ and $a_{t}$.

     \item \textbf{Value:} $\hat{q}_{t} = Q_{\theta}(z_{t},a_{t})$, predicts state-action value $\hat{q}_{t}$ given $z_{t}$ and $a_{t}$.

     \item \textbf{Policy:} $\hat{a}_{t} \sim \pi_{\theta}(z_{t})$, predicts an action $\hat{a}_{t}$ that approximately maximizes the $Q$-function.
 \end{enumerate}

 Our proposed CCEM method computes the curiosity-based intrinsic reward $r^{i}_{t}$ \textit{offline} during training using \textit{Intrinsic Curiosity Module (ICM)} \cite{icm}. As shown in Fig.\ref{fig:1}, ICM consists of an inverse dynamics model $I_{\phi}$ that takes the latent representations of two consecutive observations and predicts the action taken to move from $o_{t}$ to $o_{t+1}$, $\tilde{a}_{t} = I_{\phi}(h_{\theta}(o_{t}), h_{\theta}(o_{t+1}))$.
The inverse model is trained by minimizing the following prediction error:
\begin{equation}\label{eq:2}
    \mathcal{L}_t^\mathcal{I}(\phi,\theta_{h}) = \norm{\tilde{a}_{t} - a_{t}}^2_2.
\end{equation}

In addition to the inverse dynamics model, a forward dynamics model is required to compute the intrinsic reward as the error in predicting the next latent state. We make use of the latent dynamics $d_{\theta}$ in the TOLD model to predict the next latent state as $\hat{z}_{t+1} = d_{\theta}({z}_{t}, a_{t})$.
We follow \cite{ccfdm} in normalizing and decaying the intrinsic reward during training to converge to the optimal solutions. The intrinsic reward is computed as follows:
\begin{equation}\label{eq:3}
r^{i}_{t} = C e^{-\alpha E_t} \norm{\hat{z}_{t+1} - {z}_{t+1}}^2_2  \left( \frac{r^{max}_e}{r^{max}_i} \right),
\end{equation}
where $C$ is the intrinsic weight, $\alpha$ is the decay weight, $E_t$ is environment step, $r^{max}_e$ and $r^{max}_i$ are the maximum extrinsic and intrinsic reward respectively.
After computing the intrinsic reward, the state-action value function $Q_{\theta}$ is trained with temporal difference learning to estimate future extrinsic and intrinsic reward by minimizing the following objective:
\begin{equation}\label{eq:4}
    \mathcal{L}_t^\mathcal{Q} = \norm{Q_{\theta}(z_{t}, a_{t}) - ( r^{e}_{t} + r^{i}_{t} + \gamma Q_{\bar{\theta}}(z_{t+1},\pi_{\theta}(z_{t+1})))}^2_2,
\end{equation}
where $Q_{\bar{\theta}}$ is a target $Q$-function whose parameters are an exponential moving average (EMA) of $Q_{{\theta}}$, $\gamma$ is a discount factor, and the policy $\pi_{\theta}$ is trained to maximize $Q_{\theta}$ by minimizing the following objective:
\begin{equation}\label{eq:5}
    \mathcal{L}_t^\mathcal{\pi} = - Q_{\theta}(z_t, \pi_{\theta}(z_t)),
\end{equation}
where the policy objective is only optimized with respect to the policy parameters $\theta_{\pi}$. The reward model $R_{\theta}$ is trained by minimizing the prediction error between the predicted and the ground-truth extrinsic reward:
\begin{equation}\label{eq:6}
    \mathcal{L}_t^\mathcal{R} = \norm{R_{\theta}(z_t, a_t) - r^e_t}^2_2.
\end{equation}

In order to learn temporally predictive and consistent latent representations that are invariant to data augmentation, the subsequent observations are augmented with $\pm4$ pixel shift augmentation \cite{drq} and encoded using the target encoder $h_{\bar{\theta}}$ instead of the online encoder which is proved to be an effective practice for self-supervised representation learning \cite{SPR}, \cite{ema1}, \cite{ema2}, and a latent consistency loss is used, which is defined as follows:
\begin{equation}\label{eq:7}
    \mathcal{L}_t^\mathcal{C} = \norm{d_{\theta}(z_t, a_t) - h_{\bar{\theta}}(o_{t+1})}^2_2.
\end{equation}

Finally, the proposed TD-MPC with CCEM is trained by sampling a trajectory $\varGamma = (o_t, a_t, r^e_t, o_{t+1})_{t:t+K}$ from the replay buffer $\mathcal{B}$. Then, the TOLD model is updated by minimizing the following temporally weighted objective:
\begin{equation}\label{eq:8}
    \mathcal{L}^{TOLD}(\theta) = \sum^{t+K}_{i=t} \lambda^{i - t} (c_1\mathcal{L}_i^\mathcal{Q} + c_2\mathcal{L}_i^\mathcal{R} + c_3\mathcal{L}_i^\mathcal{C} + \mathcal{L}_i^\mathcal{\pi}),
\end{equation}
where $\lambda$ is a constant that assigns higher weight to near-term predictions, $c_1,c_2,c_3$ are loss coefficients, and $\mathcal{L}_i^\mathcal{Q},\mathcal{L}_i^\mathcal{R},\mathcal{L}_i^\mathcal{C},\mathcal{L}_i^\mathcal{\pi}$ are the single-step objectives from Eq. \ref{eq:4}, \ref{eq:6}, \ref{eq:7}, \ref{eq:5} respectively. The inverse dynamics model is updated by minimizing the following objective:
\begin{equation}\label{eq:9}
\mathcal{L}^{Inv}(\phi, \theta_{h}) = \sum^{t+K}_{i=t}\mathcal{L}_i^\mathcal{I},
\end{equation}
where $\mathcal{L}_i^\mathcal{I}$ is the single-step objective from Eq. \ref{eq:2}.

\textbf{Contrastive Learning:} To efficiently learn representations, we use contrastive learning in the form of maximizing the temporal mutual information between the joint representations of the current observation and action and the representation of the next observation  \cite{cody}. We introduce an action encoder $g_{\psi}$ that maps an action $a_{t}$ into a latent feature vector $u_t$. 
From the sampled trajectory $\varGamma_{t:t+k}$, we only use $(o_t, a_t, o_{t+1})$. The observations $o_t$ and $o_{t+1}$ are augmented and encoded using the online encoder $h_\theta$ and the target encoder $h_{\bar\theta}$ respectively:
\begin{equation} 
\begin{aligned}
z_{t} &= h_{\theta}(o_{t}), \qquad& \bar{z}_{t+1} &= h_{\bar{\theta}}(o_{t+1}).
\end{aligned}
\end{equation}

The \textit{query} is the joint representations of the current observation $o_t$ and action $a_t$ referred to as $c(z_{t}, u_t)$, where $c(., .)$ is a concatenating operation, while the representation of next observation $o_{t+1}$ referred to as $\bar{z}_{t+1}$ is the \textit{key}. 

We apply InfoNCE loss \cite{infonce} using similarity measure computed as a bilinear product $(c(z_t, u_t)^{T}W\bar{z}_{t+1})$, 
where $W$ is a learnable \textit{contrastive transformation matrix}. The temporal contrastive loss is computed as follows:
\begin{equation}\label{eq:11}
\mathcal{L}^{\mathcal{T}}(\theta_h, \psi) = - log \left[\frac{\exp(c(z_t, u_t)^{T}W\bar{z}_{t+1})}{ \sum_{\bar{z}^{j}_{t+1}\in \chi}\exp(c(z_t, u_t)^{T}W\bar{z}^{j}_{t+1})}\right],
\end{equation}
where $\chi$ is the set of all keys \textit{(positive and negative keys)}.

\subsection{Inference}

During planning, we follow TD-MPC \cite{tdmpc} except that our proposed CCEM method computes \textit{a discounted sum of Q values} over the planning horizon as the scoring function to evaluate the sampled action sequences as follows:
\begin{equation}\label{eq:10}
  \mathcal{F}_{\varGamma} =  \mathbb{E}_{\varGamma}\left[\sum^{H}_{t=0} \gamma^{t}Q_{\theta}(z_t,a_t)\right],
\end{equation}
where $\varGamma$ is a sampled action sequence, $H$ is the planning horizon, and $\gamma$ is a discount factor. Since $Q_{\theta}$ is trained by Eq. \ref{eq:4} to estimate extrinsic and intrinsic reward, CCEM encourages exploring novel states.

\RestyleAlgo{ruled}
\SetKwComment{Comment}{$\triangleright$\ }{}
\newcommand\mycommfont[1]{\footnotesize\ttfamily \textcolor{black}{#1}}
\SetCommentSty{mycommfont}
\SetKwInput{KwRequire}{Require}
\begin{algorithm}[t!]
\caption{TD-MPC with CCEM (Training)}
\label{alg:training}
\KwRequire{network parameters $(\theta, \bar{\theta}, \phi, \psi)$, replay buffer $\mathcal{B}$, learning rates $(\eta_m, \eta_i, \eta_c)$, EMA coefficeint $\zeta$, and ICM \cite{icm}}
\For{each training step}{
$\mathcal{B}\gets\mathcal{B}\cup(o_t, a_t,r^e_t, o_{t+1})_{t=0:T-1}$\Comment*[r]{Collect ep.}
\For{num updates per episode}{
$(o_t, a_t,r^e_t, o_{t+1})_{t:t+k}\sim\mathcal{B}$\Comment*[r]{Sample traj.}
$z_t = h_{\theta}(o_t)$\Comment*[r]{Encode first observation.} 
\For{$j=t$ \KwTo $t+k$}{
$\hat{r}^{e}_{j} = R_{\theta}(z_{j},a_{j})$ \Comment*[r]{ext. reward}
$\hat{q}_{j} = Q_{\theta}(z_{j},a_{j})$ \Comment*[r]{Q value} 
$\hat{a}_{j} \sim \pi_{\theta}(z_{j})$ \Comment*[r]{policy action}
$z_{j+1} = d_{\theta}(z_{j},a_{j})$\Comment*[r]{next state}
$r^i_j = ICM(o_j,a_j,o_{j+1})$ \Comment*[r]{int. reward}
}
 Update TOLD model:
$\theta\gets\theta - \eta_m \grad_{\theta} \mathcal{L}^{TOLD}(\theta)$\;
Update online encoder and inverse dynamics:
$\{\phi, \theta_h\} \gets \{\phi, \theta_h\} - \eta_i \grad_{\{\phi, \theta_h\}} \mathcal{L}^{Inv}(\phi, \theta_{h})$\;
Update online encoder and action encoder:
$\{\psi, \theta_h\} \gets \{\psi, \theta_h\} - \eta_c \grad_{\{\psi, \theta_h\}} \mathcal{L}^{\mathcal{T}}(\psi, \theta_h)$\;
Update target encoder and target $Q$-function:
$\bar{\theta} \gets (1 - \zeta)\bar{\theta} + \zeta\theta$

}
}

\end{algorithm}

\section{Experiment and Result}

\subsection{Experiment Setup}
The proposed method is evaluated on six image-based continuous control tasks from the DeepMind Control Suite \cite{dm_suite}. These tasks, shown in Fig.\ref{fig:2}, are considered a standard benchmark for evaluating image-based RL algorithms in terms of sample efficiency \cite{curl}, \cite{drq}. 

\textbf{Baselines:} We compare against previous model-based RL algorithms, such as \textit{TD-MPC} \cite{tdmpc}, \textit{PlaNet} \cite{planet}, and \textit{Dreamer} \cite{dreamer}. TD-MPC and PlaNet use real-time planning with two different variants of CEM, and Dreamer performs background planning. We also compare our method against state-of-the-art visual-based model-free RL algorithms, such as \textit{CCFDM} \cite{ccfdm}, \textit{CoDy} \cite{cody}, \textit{DrQ} \cite{drq}, and \textit{CURL} \cite{curl}. All algorithms including ours use raw images as inputs, except for \textit{SAC-State} \cite{SAC}, which is presented as an upper bound performance as it receives the direct state input from the simulator.
\subsection{Implementations}
We use the implementation of TD-MPC\footnote[1]{\scriptsize \url{ https://github.com/nicklashansen/tdmpc}} as the baseline to extend to \textit{TD-MPC with CCEM}\footnote[2]{\scriptsize \url{ https://github.com/2M-kotb/Curiosity-CEM}}. We extend the base architecture of TD-MPC by adding an inverse dynamics model $I_{\phi}$ and an action encoder $g_{\psi}$. The inverse dynamics model is implemented using a 2-layer MLP with dimension 512 and the action encoder is implemented using a 1-layer MLP with dimension 512 and all layers use ELU activations. The action encoder applies layer normalization \cite{layernorm} at the output layer and maps action into latent features vector of size 16. As observations, 3 stacked frames of $(84 \times 84)$ RGB images are used and we perform $\pm4$ pixel shift augmentation \cite{drq}. The target EMA coefficient $\zeta$ is set to $0.01$. The target $Q$-function $Q_{\bar{\theta}}$ update frequency is $2$, while the target encoder $h_{\bar{\theta}}$ update frequency is $1$. The weight constant $\lambda$ is set to $0.5$ and the loss coefficients $c_1, c_2, c_3$ are set to $0.1, 0.5$, and $2$ respectively. For the temporal contrastive loss, we find that a coefficient of $2$ gives the best performance. We use the Adam optimizer with learning rates (\num{3e-4}, \num{3e-4}, \num{1e-5}) for $(\eta_m, \eta_i, \eta_c)$ respectively, and a batch size of $256$. The intrinsic decaying weight $\alpha$ is set to \num{1e-5}. These mentioned settings are the same for all control tasks and most of the hyperparameters are adopted from TD-MPC, except those related to our method where their values are chosen heuristically. The only task-dependent hyperparameters are the intrinsic weight $C$ and the action repeat (see Table \ref{table:1}). We adopt the action repeat hyperparameters from CURL \cite{curl}.
\begin{table}[ht]
\caption{Per-task hyperparameters} 
\label{table:1}
\begin{center}
\begin{tabular}{c|c|c}
\hline
Task & Intrinsic weight$(C)$ & Action Repeat\\
\hline
Finger Spin & $0.4$ & $2$\\
Cartpole Swingup & $0.2$ & $8$\\
Reacher Easy & $0.3$ & $4$\\
Cheetah Run & $0.2$ & $4$\\
Walker Walk & $0.2$ & $2$\\
Ball-in-cup Catch & $0.3$ & $4$\\
\hline
\end{tabular}
\end{center}
\end{table}
\subsection{Experiment Results}
 \begin{table*}[ht!]
\caption{Average return (mean and standard deviations) across 5 random seeds achieved by our method and baselines on DeepMind Control Suite \cite{dm_suite} evaluated at 100k and 500k environment step. Our method outperforms model-based RL baselines by a large margin and achieves the highest average return on 4 out of 6 tasks at 100k environment step. SAC-State is an upper bound performance.} 
\label{table:2}
\centering
\begin{tabular}{c|c| c c c c|c c c c}
\multicolumn{2}{c}{} & \multicolumn{4}{c}{\textsf{\textsl{Model-free}}} & \multicolumn{4}{c}{\textsf{\textsl{Model-based}}} \\
\toprule
\multirow{2}{*}{100K step scores} & SAC-State  & CCFDM  & CoDy  & DrQ  & CURL  & PlaNet  & Dreamer  & TD-MPC  & \textbf{Ours}\\
 & \cite{SAC} & \cite{ccfdm} & \cite{cody} & \cite{drq} & \cite{cody} & \cite{ccfdm} & \cite{cody} & \cite{tdmpc} & \\
 \midrule
 Finger Spin & $672\pm76$  & $880\pm142$ & $887\pm39$ & $901\pm104$ & $750\pm37$ & $95\pm164$ & $33\pm19$ & $899\pm146$ & $\mathbf{951\pm40}$\\
 Cartpole Swingup & $812\pm45$  & $\mathbf{785\pm87}$ & $784\pm18$ & $759\pm92$ & $547\pm73$ & $303\pm71$ & $235\pm73$ & $747\pm78$ & $753\pm60$\\
 Reacher Easy & $919\pm123$ & $\mathbf{811\pm220}$ & $624\pm42$ & $601\pm213$ & $460\pm65$ & $140\pm256$ & $148\pm53$ & $413\pm62$ & $632\pm101$\\
 Cheetah Run & $228\pm95$  & $274\pm98$ & $323\pm29$ & $344\pm67$ & $266\pm27$ & $165\pm123$ & $159\pm60$ & $274\pm69$ & $\mathbf{362\pm37}$\\
 Walker Walk & $604\pm317$  & $634\pm132$ & $673\pm94$ & $612\pm164$ & $482\pm28$ & $125\pm57$ & $216\pm56$ & $653\pm99$ & $\mathbf{731\pm49}$\\
Ball-in-cup Catch & $957\pm26$  & $962\pm28$ & $948\pm6$ & $913\pm53$ & $741\pm102$ & $198\pm442$ & $172\pm96$ & $675\pm221$ & $\mathbf{964\pm3}$\\
 \midrule
500K step scores & \multicolumn{9}{c}{} \\
\midrule
 Finger Spin & $927\pm43$  & $906\pm152$ & $937\pm41$ & $938\pm103$ & $854\pm48$ & $418\pm382$ & $320\pm35$ & $\mathbf{985\pm4}$ & $980\pm9$\\
 Cartpole Swingup & $870\pm7$  & $\mathbf{875\pm38}$ & $869\pm4$ & $868\pm10$ & $837\pm15$ &$464\pm50$ & $711\pm94$ & $860\pm11$ & $864\pm6$\\
 Reacher Easy & $975\pm5$  & $\mathbf{973\pm36}$ & $957\pm16$ & $942\pm71$ & $891\pm30$ & $351\pm483$ & $581\pm160$ & $722\pm184$ & $907\pm56$\\
 Cheetah Run & $772\pm60$  & $552\pm130$ & $656\pm43$ & $\mathbf{660\pm96}$ & $492\pm22$ & $321\pm104$ & $571\pm109$ & $488\pm74$ & $531\pm49$\\
 Walker Walk & $964\pm8$  & $929\pm68$ & $943\pm17$ & $921\pm46$ & $897\pm26$ & $293\pm114$ & $924\pm35$ & $944\pm15$ & $\mathbf{946\pm6}$\\
 Ball-in-cup Catch & $979\pm6$  & $\mathbf{979\pm17}$ & $970\pm4$ & $963\pm9$ & $957\pm6$ & $352\pm467$ & $966\pm8$ & $967\pm15$ & $975\pm5$\\
\bottomrule
\end{tabular}
\end{table*}
\begin{figure*}[ht!]
\centering
     \begin{subfigure}[b]{0.49\textwidth}
         \centering
         \includegraphics[width=\textwidth]{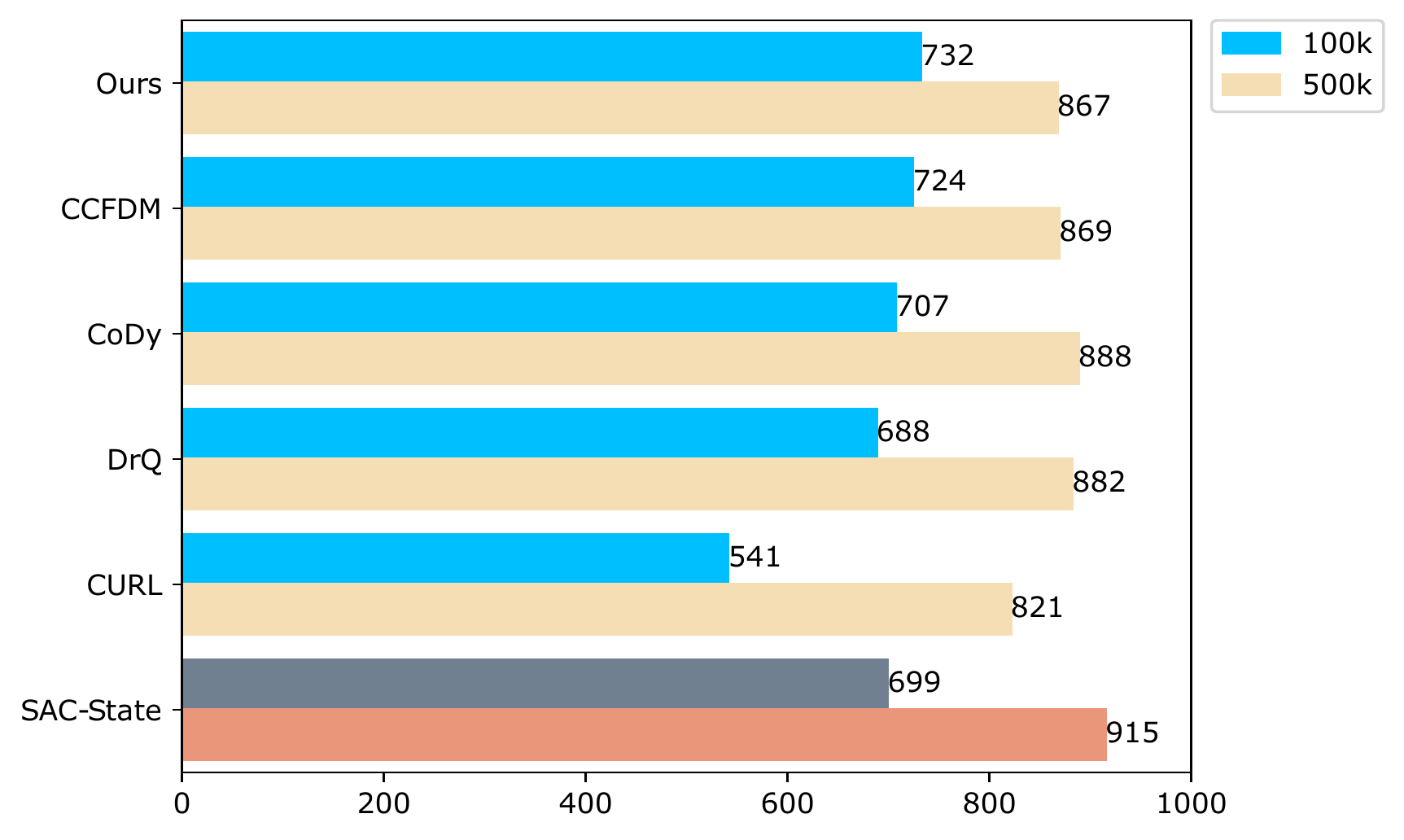}
         \caption{Ours VS. model-free RL}
         \label{fig:3a}
     \end{subfigure}
     \begin{subfigure}[b]{0.49\textwidth}
         \centering
         \includegraphics[width=\textwidth]{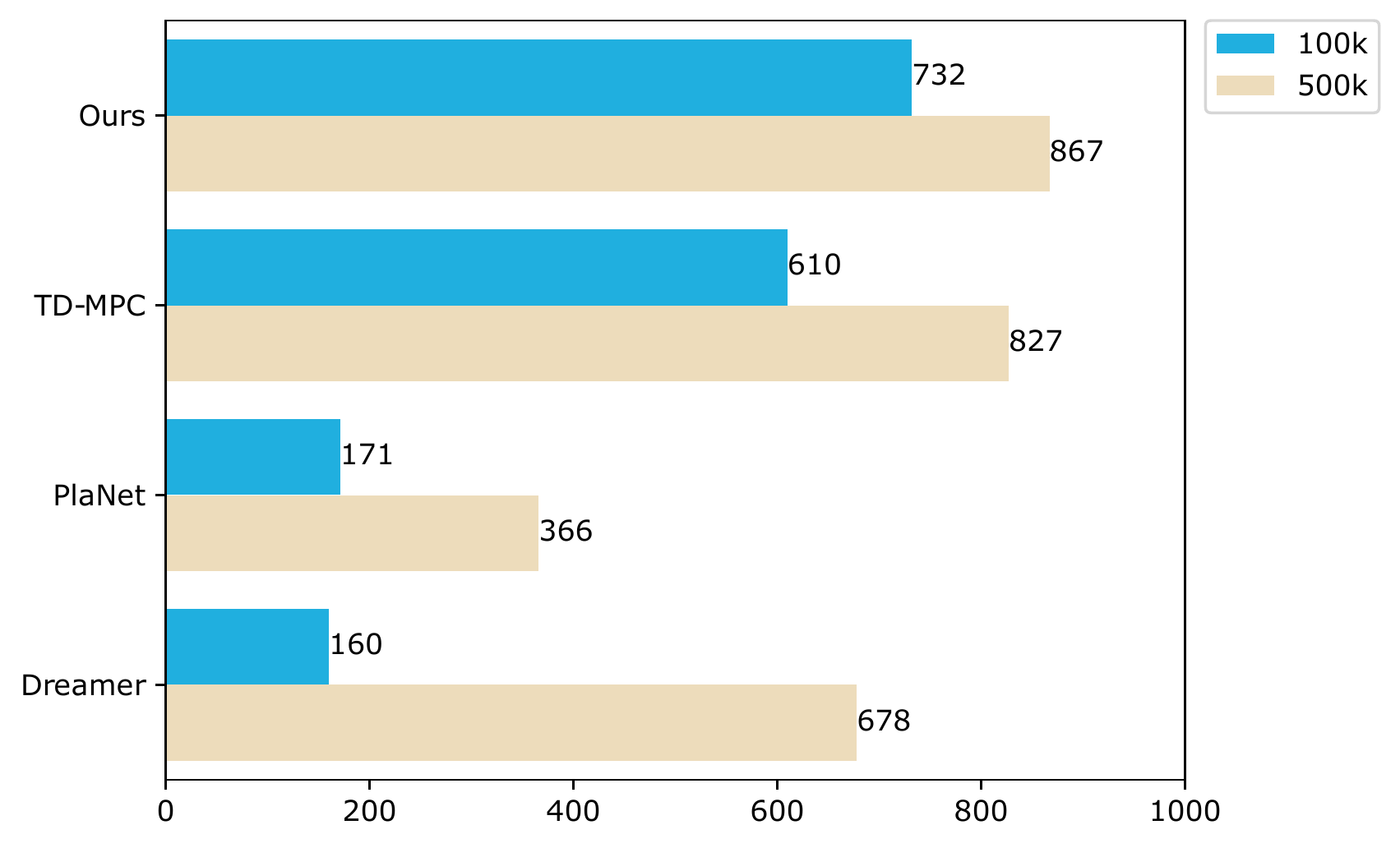}
         \caption{Ours VS. model-based RL}
         \label{fig:3b}
     \end{subfigure}
 \caption{Evaluation Score Performance of TD-MPC with CCEM averaged over six tasks relative to (a) model-free RL algorithms, (b) model-based RL algorithms. Our method outperforms all model-free baselines at 100k environment step and nearly reaches SAC-State, the upper bound performance. Our method outperforms all model-based RL baselines, as well.}
 \label{fig:3}
\end{figure*}




We run experiments for our proposed method and TD-MPC \cite{tdmpc}. For the baselines, we use the results provided in the corresponding papers, except for CURL and Dreamer, where we use the results provided in \cite{cody}, and for PlaNet, where we use the results provided in \cite{ccfdm}.
For fair comparison, we follow the settings proposed in \cite{curl}. Every agent is evaluated after every 10k environment steps, averaging over 10 episodes, then the averaged return is logged. The sample efficiency is measured by the performance at 100k environment steps, which is the relevant measure for learning speed. Also, values at 500k environment steps are given, which are near convergence. For each task, every algorithm is trained with 5 seeds and the result is reported in Table \ref{table:2}.

The result shows that TD-MPC with CCEM achieves better sample-efficiency at 100k environment steps against all baseline algorithms. TD-MPC with CCEM achieves the highest average return on four out of six tasks at 100k environment steps and close to CCFDM \cite{ccfdm} on the other two tasks (i.e., Cartpole Swingup and Reacher Easy). Our method demonstrates a stable performance with the lowest standard deviations together with CoDy \cite{cody} across all tasks which means that it is less sensitive against different seeds.

According to Fig. \ref{fig:3} that shows the average result over six tasks, our method outperforms all model-free RL baselines at 100k steps and nearly matches SAC-State \cite{SAC}, the upper bound performance  (Fig. \ref{fig:3a}), and outperforms model-based RL baselines by a large margin (Fig. \ref{fig:3b}).
TD-MPC with CCEM is the state-of-the-art model-based RL algorithm in terms of sample-efficiency which proves the robustness of CCEM as a planning method.

\subsection{Ablation Studies}

\begin{figure*}[t!]
\centering
\includegraphics[width=\textwidth]{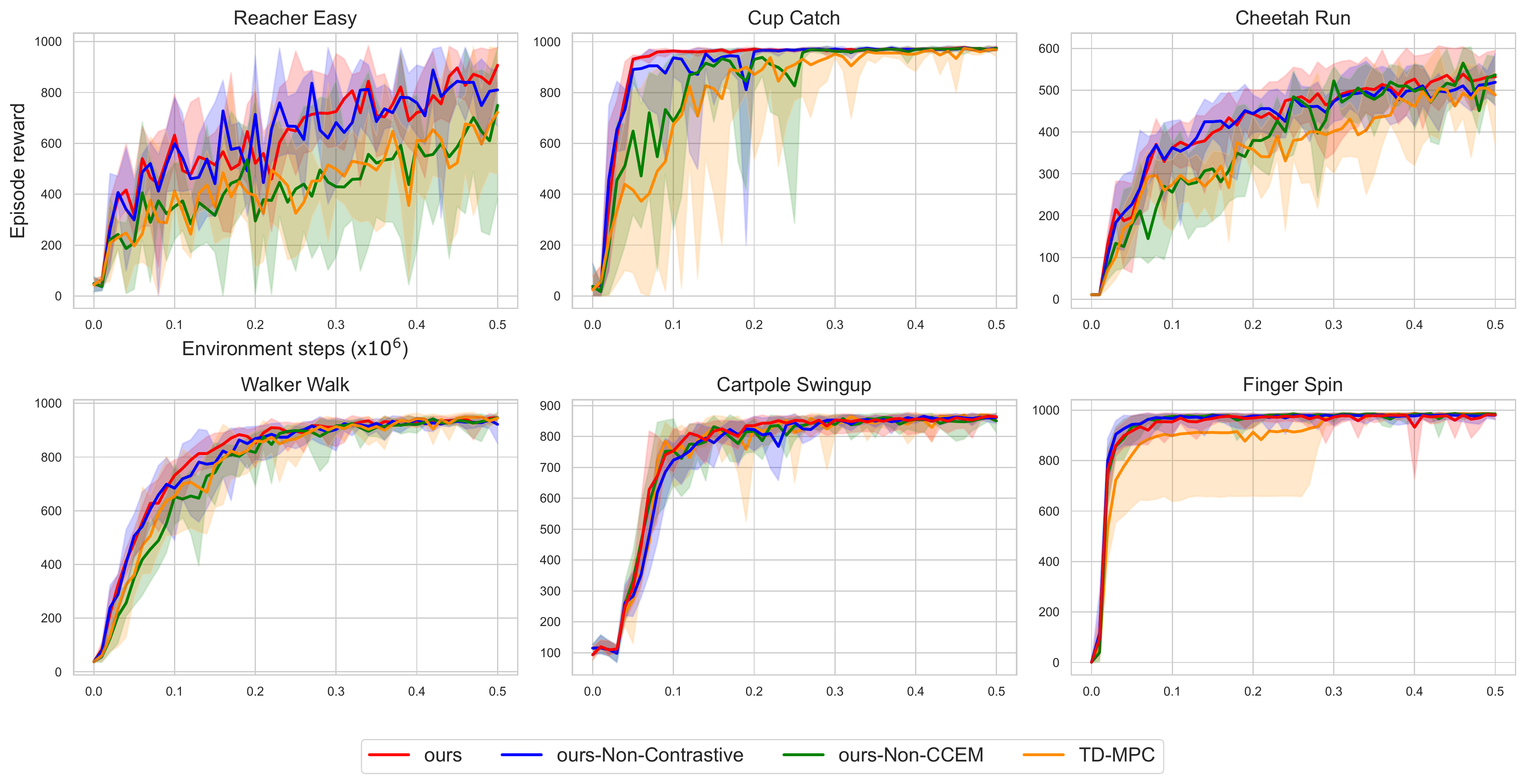}
\caption{The evaluation of the ablated variants of our method and TD-MPC as a baseline on DeepMind Control Suite.} 

\label{fig:4}
\end{figure*}

We perform ablation studies to ablate the individual contributions of our proposed Curiosity CEM planning method and contrastive representation learning. We investigate two ablations of our method: \textit{Non-Contrastive}, which is TD-MPC with the proposed CCEM planning method but without using the temporal contrastive loss, and \textit{Non-CCEM}, which is the original TD-MPC utilizing contrastive representation learning. We also include the original TD-MPC as a baseline. The evaluation of these ablations is presented in Fig. \ref{fig:4}. 

Both our method and the \textit{Non-Contrastive} variant achieve better sample-efficiency than the \textit{Non-CCEM} variant and the baseline across all tasks except for Cartpole Swingup where all the compared methods have comparable performance. Notably, the \textit{Non-CCEM} variant barely outperformed the baseline in some of the tasks such as Cup Catch, Cheetah Run and Finger Spin. On the contrary, the \textit{Non-Contrastive} variant significantly outperformed the baseline. This proves that the proposed Curiosity CEM planning method contributes the most to the success of our method while the contrastive learning barely has any contribution.
\section{Discussion and Conclusion}

In this paper, we propose the Curiosity Cross-Entropy Method (CCEM), an enhanced version of the Cross-Entropy Method for encouraging exploration via curiosity. CCEM shows that using curiosity-based intrinsic reward with the real-time planning method improves the exploration significantly and leads to a better sample-efficiency. CCEM computes the intrinsic reward \textit{offline} during training, and then learns a state-action $Q$ function to estimate extrinsic and intrinsic reward. During inference, CCEM uses a discounted sum of $Q$ values over the planning horizon as the scoring function to evaluate the sampled action sequences. 

A great advantage of our proposed planning method is that it does not increase the inference time as the computation of the intrinsic reward is done \textit{offline} during training and this highly matters in tasks that require quick responsive time such as locomotion and robotics manipulation. Furthermore, the computation of intrinsic reward is not intensive, and thus the training time is still manageable compared to other model-based RL baseline algorithms.

We select TD-MPC, a capable model-based RL algorithm to test our planning method, and we also utilize a temporal contrastive loss for better representation learning. We compared our method with state-of-the-art model-free and model-based RL algorithms on six challenging image-based benchmark tasks. The results show that our method is more sample-efficient than all the compared baselines and achieves better performance with a large margin compared to model-based RL baselines. By conducting an ablation studies on our method, we show that the proposed CCEM contributed the most to the success of our method while the contrastive learning contribution was very small and almost negligible.

CCEM proved to be a robust and sample-efficient real-time planning method and can be applied to any model-based RL algorithm as it does not require predefined conditions. For future research, we plan to evaluate CCEM with other model-based RL algorithms and test its performance with robotic manipulation tasks.

\textbf{Acknowledgement.} The authors thank Philipp Allgeuer for revising the final draft of the paper. The authors gratefully acknowledge support from the German Research Foundation DFG under project CML (TRR 169). Mostafa Kotb is funded by a scholarship from the Ministry of Higher Education of the Arab Republic of Egypt.



\bibliographystyle{IEEEtran}
\bibliography{IEEEabrv,references.bib}

\begin{thebibliography}{10}
\providecommand{\url}[1]{#1}
\csname url@rmstyle\endcsname
\providecommand{\newblock}{\relax}
\providecommand{\bibinfo}[2]{#2}
\providecommand\BIBentrySTDinterwordspacing{\spaceskip=0pt\relax}
\providecommand\BIBentryALTinterwordstretchfactor{4}
\providecommand\BIBentryALTinterwordspacing{\spaceskip=\fontdimen2\font plus
\BIBentryALTinterwordstretchfactor\fontdimen3\font minus
  \fontdimen4\font\relax}
\providecommand\BIBforeignlanguage[2]{{%
\expandafter\ifx\csname l@#1\endcsname\relax
\typeout{** WARNING: IEEEtran.bst: No hyphenation pattern has been}%
\typeout{** loaded for the language `#1'. Using the pattern for}%
\typeout{** the default language instead.}%
\else
\language=\csname l@#1\endcsname
\fi
#2}}

\bibitem{planet}
D.~Hafner, T.~Lillicrap, I.~Fischer, R.~Villegas, D.~Ha, H.~Lee, and
  J.~Davidson, ``Learning latent dynamics for planning from pixels,'' in
  \emph{International Conference on Machine Learning}.\hskip 1em plus 0.5em
  minus 0.4em\relax PMLR, 2019, pp. 2555--2565.

\bibitem{tdmpc}
N.~Hansen, X.~Wang, and H.~Su, ``Temporal difference learning for model
  predictive control,'' in \emph{International Conference on Machine Learning},
  2022.

\bibitem{dreamer}
D.~Hafner, T.~Lillicrap, J.~Ba, and M.~Norouzi, ``Dream to control: Learning
  behaviors by latent imagination,'' in \emph{International Conference on
  Learning Representations}, 2020.

\bibitem{chess}
M.~Campbell, A.~J. Hoane~Jr, and F.-h. Hsu, ``Deep blue,'' \emph{Artificial
  Intelligence}, vol. 134, no. 1-2, pp. 57--83, 2002.

\bibitem{go}
D.~Silver, A.~Huang, C.~J. Maddison, A.~Guez, L.~Sifre, G.~Van Den~Driessche,
  J.~Schrittwieser, I.~Antonoglou, V.~Panneershelvam, M.~Lanctot,
  \emph{et~al.}, ``Mastering the game of go with deep neural networks and tree
  search,'' \emph{Nature}, vol. 529, no. 7587, pp. 484--489, 2016.

\bibitem{cem}
R.~Rubinstein, ``The cross-entropy method for combinatorial and continuous
  optimization,'' \emph{Methodology and Computing in Applied Probability},
  vol.~1, pp. 127--190, 1999.

\bibitem{mbrl1}
T.~Wang and J.~Ba, ``Exploring model-based planning with policy networks,''
  \emph{arXiv preprint arXiv:1906.08649}, 2019.

\bibitem{disadvantage1}
A.~Nagabandi, G.~Kahn, R.~S. Fearing, and S.~Levine, ``Neural network dynamics
  for model-based deep reinforcement learning with model-free fine-tuning,'' in
  \emph{2018 IEEE International Conference on Robotics and Automation}.\hskip
  1em plus 0.5em minus 0.4em\relax IEEE, 2018, pp. 7559--7566.

\bibitem{mbrl2}
K.~Chua, R.~Calandra, R.~McAllister, and S.~Levine, ``Deep reinforcement
  learning in a handful of trials using probabilistic dynamics models,''
  \emph{Advances in Neural Information Processing Systems}, vol.~31, 2018.

\bibitem{icm}
D.~Pathak, P.~Agrawal, A.~A. Efros, and T.~Darrell, ``Curiosity-driven
  exploration by self-supervised prediction,'' in \emph{International
  Conference on Machine Learning}.\hskip 1em plus 0.5em minus 0.4em\relax PMLR,
  2017, pp. 2778--2787.

\bibitem{intrinsic1}
P.-Y. Oudeyer, F.~Kaplan, and V.~V. Hafner, ``Intrinsic motivation systems for
  autonomous mental development,'' \emph{IEEE Transactions on Evolutionary
  Computation}, vol.~11, no.~2, pp. 265--286, 2007.

\bibitem{intrinsic2}
Y.~Burda, H.~Edwards, D.~Pathak, A.~Storkey, T.~Darrell, and A.~A. Efros,
  ``Large-scale study of curiosity-driven learning,'' \emph{arXiv preprint
  arXiv:1808.04355}, 2018.

\bibitem{vime}
R.~Houthooft, X.~Chen, Y.~Duan, J.~Schulman, F.~De~Turck, and P.~Abbeel,
  ``{VIME}: Variational information maximizing exploration,'' \emph{Advances in
  Neural Information Processing Systems}, vol.~29, 2016.

\bibitem{intrinsic3}
S.~Mohamed and D.~Jimenez~Rezende, ``Variational information maximisation for
  intrinsically motivated reinforcement learning,'' \emph{Advances in Neural
  Information Processing Systems}, vol.~28, 2015.

\bibitem{plan2explore}
R.~Sekar, O.~Rybkin, K.~Daniilidis, P.~Abbeel, D.~Hafner, and D.~Pathak,
  ``Planning to explore via self-supervised world models,'' in
  \emph{International Conference on Machine Learning}.\hskip 1em plus 0.5em
  minus 0.4em\relax PMLR, 2020, pp. 8583--8592.

\bibitem{temporal1}
A.~Anand, E.~Racah, S.~Ozair, Y.~Bengio, M.-A. C{\^o}t{\'e}, and R.~D. Hjelm,
  ``Unsupervised state representation learning in {A}tari,'' \emph{Advances in
  Neural Information Processing Systems}, vol.~32, 2019.

\bibitem{temporal2}
K.-H. Lee, I.~Fischer, A.~Liu, Y.~Guo, H.~Lee, J.~Canny, and S.~Guadarrama,
  ``Predictive information accelerates learning in {RL},'' \emph{Advances in
  Neural Information Processing Systems}, vol.~33, pp. 11\,890--11\,901, 2020.

\bibitem{cody}
B.~You, O.~Arenz, Y.~Chen, and J.~Peters, ``Integrating contrastive learning
  with dynamic models for reinforcement learning from images,''
  \emph{Neurocomputing}, vol. 476, pp. 102--114, 2022.

\bibitem{dm_suite}
S.~Tunyasuvunakool, A.~Muldal, Y.~Doron, S.~Liu, S.~Bohez, J.~Merel, T.~Erez,
  T.~Lillicrap, N.~Heess, and Y.~Tassa, ``dm\_control: Software and tasks for
  continuous control,'' \emph{Software Impacts}, vol.~6, p. 100022, 2020.

\bibitem{shaping}
Y.~Hu, W.~Wang, H.~Jia, Y.~Wang, Y.~Chen, J.~Hao, F.~Wu, and C.~Fan, ``Learning
  to utilize shaping rewards: {{A}} new approach of reward shaping,''
  \emph{Advances in Neural Information Processing Systems}, vol.~33, pp.
  15\,931--15\,941, 2020.

\bibitem{sparse}
A.~Nair, B.~McGrew, M.~Andrychowicz, W.~Zaremba, and P.~Abbeel, ``Overcoming
  exploration in reinforcement learning with demonstrations,'' in \emph{2018
  IEEE International Conference on Robotics and Automation}.\hskip 1em plus
  0.5em minus 0.4em\relax IEEE, 2018, pp. 6292--6299.

\bibitem{count1}
M.~Bellemare, S.~Srinivasan, G.~Ostrovski, T.~Schaul, D.~Saxton, and R.~Munos,
  ``Unifying count-based exploration and intrinsic motivation,'' \emph{Advances
  in Neural Information Processing Systems}, vol.~29, 2016.

\bibitem{count2}
G.~Ostrovski, M.~G. Bellemare, A.~Oord, and R.~Munos, ``Count-based exploration
  with neural density models,'' in \emph{International Conference on Machine
  Learning}.\hskip 1em plus 0.5em minus 0.4em\relax PMLR, 2017, pp. 2721--2730.

\bibitem{count3}
M.~Lopes, T.~Lang, M.~Toussaint, and P.-Y. Oudeyer, ``Exploration in
  model-based reinforcement learning by empirically estimating learning
  progress,'' \emph{Advances in Neural Information Processing Systems},
  vol.~25, 2012.

\bibitem{ensemble1}
M.~B. Hafez, C.~Weber, M.~Kerzel, and S.~Wermter, ``Deep intrinsically
  motivated continuous actor-critic for efficient robotic visuomotor skill
  learning,'' \emph{Paladyn, Journal of Behavioral Robotics}, vol.~10, no.~1,
  pp. 14--29, 2019.

\bibitem{ensemble2}
D.~Pathak, D.~Gandhi, and A.~Gupta, ``Self-supervised exploration via
  disagreement,'' in \emph{International Conference on Machine Learning}, 2019,
  pp. 5062--5071.

\bibitem{ensemble3}
A.~Ermolov and N.~Sebe, ``Latent world models for intrinsically motivated
  exploration,'' \emph{Advances in Neural Information Processing Systems},
  vol.~33, pp. 5565--5575, 2020.

\bibitem{ensemble4}
Y.~Yao, L.~Xiao, Z.~An, W.~Zhang, and D.~Luo, ``Sample efficient reinforcement
  learning via model-ensemble exploration and exploitation,'' in \emph{2021
  IEEE International Conference on Robotics and Automation}.\hskip 1em plus
  0.5em minus 0.4em\relax IEEE, 2021, pp. 4202--4208.

\bibitem{IRRL}
M.~Li, X.~Zhao, J.~H. Lee, C.~Weber, and S.~Wermter, ``Internally rewarded
  reinforcement learning,'' \emph{arXiv preprint arXiv:2302.00270}, 2023.

\bibitem{cl1}
R.~Hadsell, S.~Chopra, and Y.~LeCun, ``Dimensionality reduction by learning an
  invariant mapping,'' in \emph{2006 IEEE Computer Society Conference on
  Computer Vision and Pattern Recognition (CVPR'06)}, vol.~2.\hskip 1em plus
  0.5em minus 0.4em\relax IEEE, 2006, pp. 1735--1742.

\bibitem{curl}
M.~Laskin, A.~Srinivas, and P.~Abbeel, ``{CURL}: Contrastive unsupervised
  representations for reinforcement learning,'' in \emph{International
  Conference on Machine Learning}.\hskip 1em plus 0.5em minus 0.4em\relax PMLR,
  2020, pp. 5639--5650.

\bibitem{infonce}
A.~v.~d. Oord, Y.~Li, and O.~Vinyals, ``Representation learning with
  contrastive predictive coding,'' \emph{arXiv preprint arXiv:1807.03748},
  2018.

\bibitem{atc}
A.~Stooke, K.~Lee, P.~Abbeel, and M.~Laskin, ``Decoupling representation
  learning from reinforcement learning,'' in \emph{International Conference on
  Machine Learning}.\hskip 1em plus 0.5em minus 0.4em\relax PMLR, 2021, pp.
  9870--9879.

\bibitem{SPR}
M.~Schwarzer, A.~Anand, R.~Goel, R.~D. Hjelm, A.~C. Courville, and P.~Bachman,
  ``Data-efficient reinforcement learning with self-predictive
  representations,'' in \emph{International Conference on Learning
  Representations}, 2021.

\bibitem{ccfdm}
T.~Nguyen, T.~M. Luu, T.~Vu, and C.~D. Yoo, ``Sample-efficient reinforcement
  learning representation learning with curiosity contrastive forward dynamics
  model,'' in \emph{2021 IEEE/RSJ International Conference on Intelligent
  Robots and Systems}.\hskip 1em plus 0.5em minus 0.4em\relax IEEE, 2021, pp.
  3471--3477.

\bibitem{dreaming}
M.~Okada and T.~Taniguchi, ``Dreaming: Model-based reinforcement learning by
  latent imagination without reconstruction,'' in \emph{2021 IEEE International
  Conference on Robotics and Automation}.\hskip 1em plus 0.5em minus
  0.4em\relax IEEE, 2021, pp. 4209--4215.

\bibitem{cvrl}
X.~Ma, S.~Chen, D.~Hsu, and W.~S. Lee, ``Contrastive variational reinforcement
  learning for complex observations,'' \emph{arXiv preprint arXiv:2008.02430},
  2020.

\bibitem{tpc}
T.~D. Nguyen, R.~Shu, T.~Pham, H.~Bui, and S.~Ermon, ``Temporal predictive
  coding for model-based planning in latent space,'' in \emph{International
  Conference on Machine Learning}.\hskip 1em plus 0.5em minus 0.4em\relax PMLR,
  2021, pp. 8130--8139.

\bibitem{dreamerpro}
F.~Deng, I.~Jang, and S.~Ahn, ``Dreamerpro: Reconstruction-free model-based
  reinforcement learning with prototypical representations,'' in
  \emph{International Conference on Machine Learning}.\hskip 1em plus 0.5em
  minus 0.4em\relax PMLR, 2022, pp. 4956--4975.

\bibitem{drq}
D.~Yarats, I.~Kostrikov, and R.~Fergus, ``Image augmentation is all you need:
  Regularizing deep reinforcement learning from pixels,'' in
  \emph{International conference on learning representations}, 2020.

\bibitem{terminal_value}
V.~Feinberg, A.~Wan, I.~Stoica, M.~I. Jordan, J.~E. Gonzalez, and S.~Levine,
  ``Model-based value expansion for efficient model-free reinforcement
  learning,'' in \emph{Proceedings of the 35th International Conference on
  Machine Learning}, 2018.

\bibitem{value-sum}
M.~Raisi, A.~Noohian, L.~Mccutcheon, and S.~Fallah, ``Value {S}ummation: {A}
  novel scoring function for {MPC}-based model-based reinforcement learning,''
  \emph{arXiv preprint arXiv:2209.08169}, 2022.

\bibitem{ema1}
A.~Tarvainen and H.~Valpola, ``Mean teachers are better role models:
  Weight-averaged consistency targets improve semi-supervised deep learning
  results,'' \emph{Advances in Neural Information Processing Systems}, vol.~30,
  2017.

\bibitem{ema2}
J.-B. Grill, F.~Strub, F.~Altch{\'e}, C.~Tallec, P.~Richemond, E.~Buchatskaya,
  C.~Doersch, B.~Avila~Pires, Z.~Guo, M.~Gheshlaghi~Azar, \emph{et~al.},
  ``Bootstrap your own latent {A} new approach to self-supervised learning,''
  \emph{Advances in Neural Information Processing Systems}, vol.~33, pp.
  21\,271--21\,284, 2020.

\bibitem{SAC}
T.~Haarnoja, A.~Zhou, P.~Abbeel, and S.~Levine, ``Soft actor-critic: Off-policy
  maximum entropy deep reinforcement learning with a stochastic actor,'' in
  \emph{International Conference on Machine Learning}.\hskip 1em plus 0.5em
  minus 0.4em\relax PMLR, 2018, pp. 1861--1870.

\bibitem{layernorm}
J.~L. Ba, J.~R. Kiros, and G.~E. Hinton, ``Layer normalization,'' \emph{arXiv
  preprint arXiv:1607.06450}, 2016.

\end{thebibliography}

\end{document}